\documentclass[fp]{jpsj3}

\usepackage{txfonts}
\usepackage{amsmath}
\usepackage{bm}
\usepackage{graphicx}
\usepackage{algorithm}
\usepackage{algpseudocode}
\usepackage{booktabs}
\usepackage{dsfont}
\usepackage{siunitx}
\usepackage{url}

\newcommand{\argmin}{\mathop{\mathrm{arg\,min}}}

\newcommand{\IG}{\mathrm{IG}}
\newcommand{\R}{\mathbb{R}}

\title{Robust Wavelength Selection for Partial Least Squares Sugar Content Estimation Using Combinatorial Bayesian Optimization}

\author{Mitsunobu~Kanebako$^1$, Ami~S.~Koshikawa$^1$, Masaru~Hitomi$^1$, Takuro~Tanaka$^2$, Mahito~Chiba$^2$, Maiko~Mori$^2$, and Masayuki~Ohzeki$^{1,3,4,5}$}

\inst{
$^1$Graduate School of Information Sciences, Tohoku University, Miyagi 980-8564, Japan \\
$^2$DIC Corporation, Tokyo 103-8233, Japan \\
$^3$Department of Physics, Institute of Science Tokyo, Tokyo 152-8551, Japan \\
$^4$Research and Education Institute for Semiconductors and Informatics, Kumamoto University, Kumamoto 860-0862, Japan \\
$^5$Sigma-i Co., Ltd., Tokyo 108-0075, Japan
}

\abst{
Wavelength selection is one of the important preprocessing methods in near-infrared spectroscopy
to improve prediction accuracy and interpretability of spectral data.
We formulate wavelength-region selection for sugar content estimation as a binary black-box optimization problem
and propose a method based on Bayesian optimization.
The proposed method constructs a sparse quadratic surrogate model
and sequentially extracts interested wavelength regions by Thompson sampling.
Minimizing an acquisition function is performed as a quadratic unconstrained binary optimization problem by simulated or quantum annealing.
Experiments show that the proposed method improves the prediction accuracy of partial least squares regression
and yields more consistent wavelength regions than genetic-algorithm-based selection and simulated annealing.
Under one-bit local perturbations, the selected wavelength regions show minimal fluctuations in root mean square errors between observed and predicted values of a validation set.
This local stability suggests that our method converges to a smoother error landscape and avoids isolated overfitted solutions.
These results indicate that combinatorial Bayesian optimization is a useful framework for robust feature selection in spectroscopic prediction tasks.
}

\begin{document}
\maketitle

\section{Introduction}

Near-infrared spectroscopy is a nondestructive and rapid analytical technique for estimating physical and chemical properties of samples.
It has been used in a wide range of applications, including quality evaluation of agricultural products\cite{guthrieNIRModelDevelopment2006} and biomedical sensing\cite{koguchiDevelopmentClinicalApplication2002}.
Particularly, in fruit-quality assessment, near-infrared spectra provide useful information for estimating internal properties such as sugar content and ripeness.

A practical difficulty in near-infrared spectral analysis\cite{becInterpretabilityNIR2025} includes high-dimensionality and multicollinearity of near-infrared spectra.
A single spectrum often contains hundreds or thousands of wavelength points, many of which are linearly dependent.
Such multicollinearity can introduce unnecessary degrees of freedom into regression models and reduce the interpretability of the resulting explanatory variables.
In addition, measurement noise, environmental variation, and sample-to-sample variability unrelated to target materials, such as scattering effects, can introduce irrelevant components, which may lead to overfitting and poor generalization.
Although high-dimensional machine-learning models can improve prediction accuracy and partly mitigate the influence of noise and overfitting\cite{niNonlinearCalibrationModels2014}, their models often reduce interpretability of which spectral regions are responsible for the prediction.
Wavelength selection is, therefore, an important approach for identifying informative spectral regions and improving generalization performance.

Wavelength selection often involves black-box optimization, which is designed to optimize an expensive-to-evaluate function, to automatically identify regions of interest within spectra.
Previous research proposed a selection method that utilized a genetic algorithm (GA) with partial least squares regression\cite{kawamuraDevelopmentGeneticAlgorithmbased2006}, and it allows us to search over a large combinatorial space of candidate wavelength selection patterns.
Since being a population-based method, this method tends to avoid a specific local minimum.
At the same time, it is more likely to transition between very different selection patterns but similar corresponding objective values.
This variability makes it difficult to identify whether a selected wavelength region is intrinsically important or merely selected by chance.
Simulated annealing (SA)\cite{kirkpatrickOptimizationSimulatedAnnealing1983} can also be applied to the same binary selection problem, but it requires repeated evaluations of the expensive objective function and can still produce inconsistent solutions.

The objective of this study is to develop a more sample-efficient and reproducible wavelength-selection method for near-infrared spectral prediction.
We formulate the selection of wavelength regions as a binary black-box optimization problem and apply Bayesian Optimization of Combinatorial Structures (BOCS)\cite{baptistaBayesianOptimizationCombinatorial2018}.
The method fits a sparse quadratic surrogate model to previously evaluated wavelength selection patterns and proposes the next wavelength selection by Thompson sampling.
The resulting acquisition function in Thompson sampling is formed as a quadratic unconstrained binary optimization model, which can be handled by classical SA or quantum annealing (QA).

QA\cite{kadowakiQuantumAnnealingTransverse1998} is a hardware-based metaheuristic for binary optimization problems that uses quantum dynamics to search for low-energy states, 
whereas SA uses thermal fluctuations. 
Because the acquisition function in BOCS is expressed as a quadratic unconstrained binary optimization problem, 
QA can be used as an alternative optimizer of the acquisition function. 
Previous studies have applied QA to a variety of combinatorial optimization and black-box optimization problems
\cite{Tanaka2023,Doi2023,rosenberg2016solving,venturelli2019reverse,orusForecastingFinancialCrashes2019,mugel_dynamic_2022,neukart2017traffic,shikanai2023,feldHybridSolutionMethod2019,sawamuraQuantumclassicalHybridAlgorithm2025,ohzekiControlAutomatedGuided2019,Haba2022,Yonaga2022,nishimura_item_2019}, 
suggesting its potential as a hardware-compatible solver for large binary search spaces. 
In this study, however, QA is used only to optimize the BOCS acquisition function, not to evaluate the PLS objective itself. 
We therefore regard QA as one candidate optimizer for Thompson sampling and compare its performance with that of SA within the same BOCS framework.

This study has three main contributions.
First, we formulate partial-least-squares-based wavelength selection as a binary black-box optimization problem whose objective is the cross-validated prediction error.
Second, we apply combinatorial Bayesian optimization with a sparse quadratic surrogate model to this problem.
Third, we evaluate not only prediction accuracy but also the consistency of selected wavelength regions over repeated runs and the robustness of the obtained solution against local perturbations.
This evaluation is important because a useful wavelength-selection method should provide stable spectral regions that can be examined by domain experts.

The rest of this paper is organized as follows.
Section~\ref{sec_problem} defines the optimization problem and describes the proposed method.
Section~\ref{sec_exp} explains the dataset, preprocessing, baseline methods, and evaluation metrics.
Section~\ref{sec_res} presents the experimental results and discusses the implications and limitations of the method.
We summarize the conclusions in Sect.~\ref{sec_con}.

\section{Problem Formulation and Method}
\label{sec_problem}

\subsection{Wavelength selection}

Suppose that a near-infrared spectrum has a certain resolution $\Delta\lambda$ and we handle the whole wavelength regions as a collection of $n$ small regions with the width of $\Delta\lambda$.
We represent a candidate selection of wavelength regions by a binary vector
$\bm{x}=(x_1,\ldots,x_n)\in\{0,1\}^n$,
where $x_i=1$ indicates that the $i$th wavelength region is selected and $x_i=0$ does that it is excluded.
Each intensity at each measured wavelength point is stored in a matrix $I\in \R^{N\times n}$, where $N$ represents the number of spectra.
We extract selected columns $i$ from $I$ that satisfy $x_i = 1$ and define the submatrix as $I_{\bm{x}} \in \R^{N\times\|x\|_0}$.

For a given binary vector $\bm x$, the prediction error is evaluated by partial least squares regression with $K$-fold cross validation.
Because the number of latent components strongly affects the prediction performance, the number of components is selected within the objective evaluation.
We define $m_{\bm x}$
as the maximum number of components considered.
For the $i$-th fold and the number of components $r$,
the root mean square error (RMSE) between the validation set of observation $y^{(i)}_\mathrm{val}$ and prediction $\hat{y}^{(i)}_\mathrm{val}$ is
\begin{equation}
  R_i(r,\bm x)=
  \sqrt{
  \frac{1}{N_{\mathrm{val}}^{(i)}}
  \left\|
  \hat{y}_{\mathrm{val}}^{(i)}(r,\bm x)
  -
  y_{\mathrm{val}}^{(i)}
  \right\|_2^2
  },
\end{equation}
where $N_\mathrm{val}^{(i)}$ is the number of elements in a vector $y_\mathrm{val}$.
The cross-validated error for $r$ components is then
\begin{equation}
  R_\mathrm{CV}(r,\bm x)=\frac{1}{K}\sum_{i=1}^{K}R_i(r,\bm x).
\end{equation}
The black-box objective function used in this study is
\begin{equation}
  L(\bm x)=\min_{1\le r\le m_{\bm x}}R_\mathrm{CV}(r,\bm x).
  \label{eq:objective}
\end{equation}
The wavelength-selection problem is therefore written as
\begin{equation}
  \bm x^*=\argmin_{\bm x\in\{0,1\}^n} L(\bm x).
  \label{eq:binary_problem}
\end{equation}
We handle $L(\bm x)$ as a black-box objective function and solve the wavelength-region selection task as a combinatorial optimization problem over binary selection vectors.

\begin{algorithm}[t]
\caption{Objective evaluation $L(\bm x)$}
\label{alg:observe_pls}
\begin{algorithmic}[1]
\Require Selection vector $\bm x\in\{0,1\}^n$, spectral matrix $I\in\R^{N\times n}$, target vector $\bm y\in\R^N$, and number of folds $K$.
\State Extract $I_{\bm x}\in \R^{N\times\|\bm x\|_0}$ from $I$ using columns with $x_i=1$.
\State Set $m_{\bm x}$.
\State Set $L_{\mathrm{best}}\gets\infty$.
\For{$r=1$ to $m_{\bm x}$}
  \State Set $L_r\gets0$.
  \For{$i=1$ to $K$}
    \State Train a PLS model with $r$ components using all folds except fold $i$.
    \State Predict the target values for fold $i$.
    \State Add the fold RMSE to $L_r$.
  \EndFor
  \State Set $L_r\gets L_r/K$.
  \If{$L_r<L_{\mathrm{best}}$}
    \State Set $L_{\mathrm{best}}\gets L_r$.
  \EndIf
\EndFor
\State \Return $L_{\mathrm{best}}$.
\end{algorithmic}
\end{algorithm}

\subsection{Sparse quadratic surrogate model}

Combinatorial Bayesian optimization\cite{baptistaBayesianOptimizationCombinatorial2018} approximates the objective function to a surrogate model using previously evaluated binary vectors. At iteration $t$, the dataset is
\begin{equation}
  \mathcal D_t=\{(\bm x^{(q)},L(\bm x^{(q)}))\}_{q=1}^{t}.
\end{equation}
We use a quadratic surrogate model,
\begin{equation}
  \hat f(\bm x;\bm\alpha)=\alpha_0+\sum_{i=1}^{n}\alpha_i x_i+
  \sum_{1\le i<j\le n}\alpha_{ij}x_i x_j,
  \label{eq:quadratic_surrogate}
\end{equation}
where $\bm\alpha$ denotes all regression coefficients. This model contains one-body and two-body terms and can express pairwise interactions among wavelength regions while keeping the model in the quadratic unconstrained binary optimization form.

The posterior distribution of the coefficients is written as
\begin{equation}
  p(\bm\alpha|\mathcal D_t)\propto p(\mathcal D_t|\bm\alpha)p(\bm\alpha).
\end{equation}
A horseshoe prior\cite{carvalhoHorseshoeEstimatorSparse2010} is used for $p(\bm\alpha)$ to encourage sparsity. This prior shrinks most coefficients toward zero while allowing a small number of coefficients to remain large. This property is suitable for wavelength selection, where only a limited number of spectral regions are expected to contribute substantially to prediction.

Here we assume the likelihood function to follow a Gaussian distribution $\mathcal{N}(X\bm \alpha,\, \sigma^2 \mathds{1}_t)$, where $X\in\{0,\,1\}^{t\times p}\ (p=1 + n + n(n-1)/2)$, $\sigma^2$, and $\mathds{1}_t$ represent the design matrix for the quadratic model, the noise variance, and the identity matrix with the size $t$, respectively.
Following the standard Gibbs-sampling implementation of the horseshoe prior, the conditional distribution of $\bm\alpha$ and $\sigma^2$ can be expressed\cite{makalicSimpleSamplerHorseshoe2016} as
\begin{align}
  \bm\alpha|\cdot &\sim N(A^{-1}X^T\bm\ell,\sigma^2 A^{-1}),\label{eq:gibbs_alpha}\\
  A&=X^TX+\Sigma_*^{-1}, \nonumber \\
  \Sigma_*&=\tau^2\mathrm{diag}(\beta_1^2,\ldots,\beta_p^2), \nonumber \\
  \sigma^2|\cdot &\sim \IG\left(\frac{t+p}{2},
  \frac{(\bm\ell-X\bm\alpha)^T(\bm\ell-X\bm\alpha)}{2}
  +\frac{\bm\alpha^T\Sigma_*^{-1}\bm\alpha}{2}\right), \nonumber \\
  \beta_k^2|\cdot &\sim \text{IG}\left(1, \frac{1}{\nu_k} + \frac{\alpha_k^2}{2\tau^2\sigma^2}\right), \quad k = 1, \ldots, p \nonumber \\
\tau^2|\cdot &\sim \text{IG}\left(\frac{p+1}{2}, \frac{1}{\xi} + \frac{1}{2\sigma^2}\sum_{k=1}^{p}\frac{\alpha_k^2}{\beta_k^2}\right) \nonumber \\
\nu_k|\cdot &\sim \text{IG}\left(1, 1 + \frac{1}{\beta_k^2}\right), \quad k = 1, \ldots, p \nonumber \\
\xi|\cdot &\sim \text{IG}\left(1, 1 + \frac{1}{\tau^2}\right) \nonumber
\end{align}
where $\bm\ell\in\R^t$ denotes the vector of observed objective values.
The local scales $\beta_k^2$, global scale $\tau^2$, and auxiliary variables are sampled from the corresponding inverse-gamma conditional distributions. In practice, this posterior sampling step is the main overhead ($\mathcal{O}(t^2p)$)\cite{bhattacharyaFastSamplingGaussian2016} of the proposed method when $t$ is large, whereas the expensive objective evaluation in Eq.~(\ref{eq:objective}) is performed only once per newly proposed binary vector.

\subsection{Thompson sampling and quadratic acquisition function}

At each iteration, a parameter vector $\bm\alpha_t$ is sampled from the posterior distribution.
The next candidate is selected by optimizing the sampled surrogate model,
\begin{equation}
  \bm x_{t+1}=\argmin_{\bm x\in\{0,1\}^n}\hat f(\bm x;\bm\alpha_t),
  \quad
  \bm\alpha_t\sim p(\bm\alpha|\mathcal D_t).
  \label{eq:ts}
\end{equation}
This is Thompson sampling for minimization. Since $\hat f$ is quadratic in binary variables, Eq.~(\ref{eq:ts}) is a quadratic unconstrained binary optimization problem,
\begin{equation}
  E_t(\bm x)=\alpha_0^{(t)}+\sum_i\alpha_i^{(t)}x_i+
  \sum_{i<j}\alpha_{ij}^{(t)}x_ix_j.
  \label{eq:qubo}
\end{equation}
The constant term does not affect the minimizer. We optimize this acquisition function by SA or QA.

\begin{algorithm}[t]
\caption{BOCS-based wavelength selection}
\label{alg:bocs}
\begin{algorithmic}[1]
\Require Initial dataset $\mathcal D_{t_0}$ and total number of evaluations $T$.
\For{$t=t_0$ to $T-1$}
  \State Sample $\bm\alpha_t\sim p(\bm\alpha|\mathcal D_t)$.
  \State Construct the quadratic energy $E_t(\bm x)$ in Eq.~(\ref{eq:qubo}).
  \State Obtain $\bm x_{t+1}\approx\argmin_{\bm x\in\{0,1\}^n}E_t(\bm x)$ by either SA or QA.
  \State Evaluate $L(\bm x_{t+1})$.
  \State Update $\mathcal D_{t+1}\gets\mathcal D_t\cup\{(\bm x_{t+1},L(\bm x_{t+1}))\}$.
\EndFor
\State \Return $\argmin_{(\bm x,L)\in\mathcal D_T}L$.
\end{algorithmic}
\end{algorithm}

\subsection{Annealing algorithms}

SA is a classical heuristic for minimizing an energy function. Given a current state $\bm x$ and a proposed state $\bm x'$, the transition is accepted with probability
\begin{equation}
P(\bm x'|\bm x)=
\begin{cases}
    1 & \textrm{if}\ E(\bm x')\le E(\bm x),\\
    \exp\left[-\{E(\bm x')-E(\bm x)\}/T\right] & \textrm{otherwise},
\end{cases}
\end{equation}
where $T$ is the temperature. The temperature is gradually reduced during the search, and one-spin flips are used as elementary proposals.

QA, on the other hand, represents a binary optimization problem by a problem Hamiltonian and searches for a low-energy state through quantum dynamics.
The problem Hamiltonian is often formulated as a two-body Ising model:
\begin{equation}
  H_P=\sum_i h_i\sigma_i+\sum_{i<j}J_{ij}\sigma_i\sigma_j,
\end{equation}
where $\sigma_i = \{-1,\,1\}$.
The quadratic unconstrained binary optimization (QUBO) form in Eq.~(\ref{eq:qubo}) can be converted into this Ising form by the standard transformation between binary variables and spin variables. When a hardware quantum annealer is used, minor embedding and chain-strength tuning are required because the logical connectivity of a dense QUBO generally differs from the hardware graph\cite{dattaniPegasusSecondConnectivity2019,boothbyNextGenerationTopologyDWave2020}.
In the present study, QA is used only to optimize the acquisition function in BOCS, not to evaluate the PLS objective itself.

\section{Experimental Setup}
\label{sec_exp}

\subsection{Dataset and preprocessing}

We used $40$ pairs of near-infrared spectra of plums and their measured sugar contents in degrees Brix ($^\circ\mathrm{Bx}$)\cite{nirplums2024} and
they were centered around approximately $20\,^\circ$Bx.
Each raw spectrum contained 600 wavelength points. 
Before wavelength selection, multiplicative scatter correction was applied to reduce multiplicative and additive scattering effects.  
For a measured spectrum $\bm s_i$, the corrected spectrum was computed as
\begin{equation}
  \bm s_i^{\mathrm{MSC}}=\frac{\bm s_i-b_i}{a_i},
\end{equation}
where $a_i$ and $b_i$ were obtained by linearly regressing $\bm s_i$ against the mean spectrum.

The corrected spectra were divided into $n$ contiguous wavelength regions.  The average intensity in each region was used as a feature, giving an $n$-dimensional vector for each sample.  The number of regions $n$ controls the resolution of wavelength selection.  Larger $n$ allows a finer spectral representation but increases the dimension of the binary optimization problem, whereas smaller $n$ reduces the search space but can smooth out localized spectral information.  We tested $n\in\{20,30,40,60,75,100\}$ and selected the value used in the main comparison from the prediction results.
In addition, to smooth the spectra and remove constant offsets, we used the Savitzky-Golay filter\cite{savitzkySmoothingDifferentiationData1964} and then obtained the first derivative of the absorption spectra.

\subsection{Compared wavelength-selection methods}

We compared the proposed BOCS-based method with GA-based wavelength selection and SA.  All methods generated binary selection vectors and used the same objective evaluation $L(\bm x)$.
PLS regression in the evaluation of $L(\bm x)$ requires the number of latent variables as a hyperparameter.
Here, we defined the maximum number of latent variables $m_{\bm x}$ as
\begin{align}
    m_{\bm x} = \min \{\|\bm x\|_0, 35\},
\end{align}
and searched for the number that minimized $L(\bm x)$ by brute force from 1 to $m_{\bm x}$.

\subsubsection{Genetic algorithm}

In the GA baseline, each individual represents a binary wavelength-selection vector.  The fitness of an individual is evaluated using $L(\bm x)$.  Selection, crossover, and mutation are repeated over generations to search for a low-error subset.  The number of generations was 500, the population size was 60, the crossover probability was 1, the mutation probability was 0.01 per bit, and the survival rate was 0.5.

\subsubsection{Simulated annealing}

In the SA baseline, the objective function $L(\bm x)$ was evaluated directly for candidate wavelength subsets.  Neighboring states were generated by one-spin flips, $x_i:0\leftrightarrow1$, and the transition was accepted or rejected according to the Metropolis rule.  The initial temperature was $T_0=10$, the cooling rate was $0.98$, and the number of iterations was 500.  We also examined a random one-flip variant to compare the effect of the number of objective evaluations.

\subsubsection{BOCS-based method}

For the proposed method, the initial dataset was generated by random binary vectors.
At each subsequent iteration, the quadratic surrogate model was fitted to the accumulated observations, and the acquisition function was minimized by either SA or QA.
For SA, to optimize the acquisition function, we used the OpenJij SA sampler with 100 reads and 100 sweeps.
For QA, we used the \texttt{advantage\_system4.1} quantum processing unit\cite{McGeochAdvantageSystem2021} provided from D-Wave Systems Inc. with an annealing time of $\SI{20}{\micro s}$.
Mapping the fully-connected graph onto the sparse graph of the quantum processor, called minor embedding, was achieved by the heuristic \texttt{minorminer}\cite{boothbyFastCliqueMinor2020}.
After minor embedding, multiple qubits on the hardware graph can represent a single bit on a problem graph.
These qubits are, however, unlikely to have identical values, which is called chain breaking. To handle broken chains, we used a majority voting and unembeded solutions from the hardware.
We sampled 10 solutions at each iteration and selected one that realized the smallest value of the acquisition function.
In both cases, the expensive PLS-based objective was evaluated only after a new candidate wavelength subset had been proposed.

\subsection{Evaluation metrics}

The methods were evaluated from three viewpoints: prediction accuracy, consistency of selected wavelength subsets, and local robustness of the obtained solution.

Prediction accuracy was measured by the cross-validated RMSE returned by $L(\bm x)$.  We compared the average behavior over repeated runs and the average of the best values obtained within the specified number of evaluations from independent trials.

Consistency was quantified by the mean Hamming distance among selected binary vectors.  Let $\bm x^{(r)}\in\{0,1\}^n$ be the selected vector obtained in the $r$th run, where $r=1,\ldots,R$.  The mean Hamming distance is
\begin{equation}
  \bar d_H=\frac{2}{R(R-1)}\sum_{1\le r<s\le R}
  d_H\left(\bm x^{(r)},\bm x^{(s)}\right),
  \label{eq:hamming_mean}
\end{equation}
where
\begin{equation}
  d_H(\bm a,\bm b)=\sum_{i=1}^{n}{\bf 1}(a_i\ne b_i).
\end{equation}
A smaller $\bar d_H$ means that repeated runs select more similar wavelength subsets.

Local robustness was evaluated by applying one-bit perturbations to a representative selected vector.  For each selected component with $x_i=1$, we changed it to $x_i=0$ and measured the increase in RMSE.  If the RMSE remains stable under most one-bit removals, the solution is interpreted as robust to local perturbations.  Conversely, if many removals lead to large RMSE increases, the selected subset is sensitive to small changes.

\section{Results and Discussion}
\label{sec_res}

\subsection{Effect of the number of wavelength regions}

\begin{figure}[t]
  \centering
  \includegraphics[width=.9\linewidth]{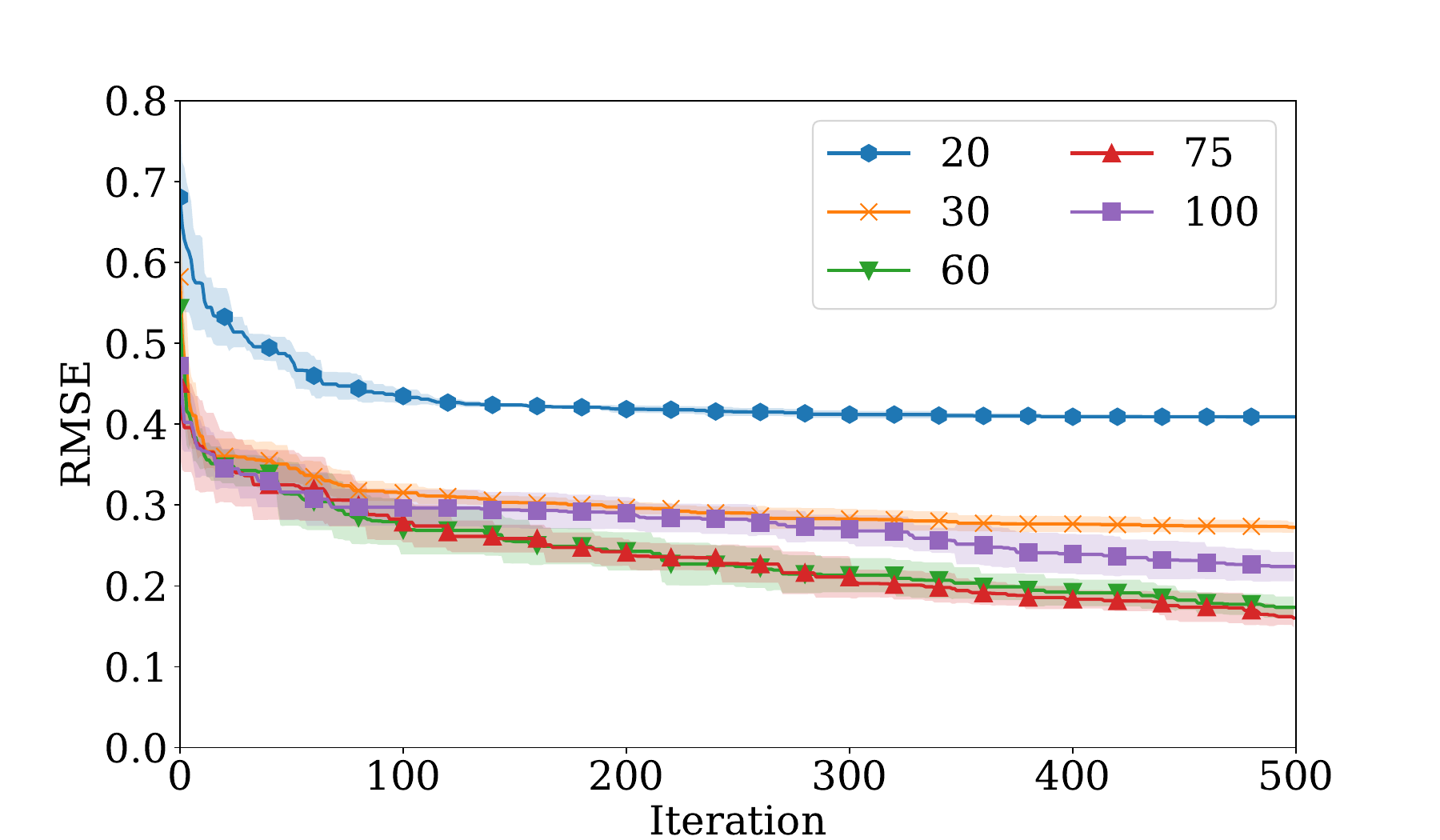}
    \caption{(Color online) RMSE trajectories of BOCS for different numbers of wavelength regions. The solid lines and the shaded areas represent the mean values and standard deviations of trials, respectively. $n=60$ and $75$ curves achieve smaller RMSE than any other after 500 iterations.}
  \label{fig_RMSE_n}
\end{figure}

Figure~\ref{fig_RMSE_n} shows the RMSE trajectories for different numbers of wavelength regions.  
The RMSE after 500 iterations was small for $n=60$ and $n=75$.  
For larger values of $n$, the RMSE was still decreasing at the end of 500 iterations, 
suggesting that more iterations would be required to search the larger binary space. 
For $n=20$ and $30$, 
the wavelength region corresponding to each segment becomes relatively broad, and the spectral values within each segment are averaged.  
As a result, the complex structure contained in the original spectra may no longer be sufficiently captured, which likely caused the RMSE to stop decreasing during the optimization.  
In contrast, 
for a finer division such as $n=100$, 
the RMSE was still decreasing at the end of 500 iterations.
This behavior is presumably due to the increased dimensionality of the explanatory variable $\bm{x}$, which expands the search space and therefore requires a larger number of trials to find combinations of $\bm{x}$ that yield a smaller RMSE.  
Since practical applications require the optimization to be terminated after a finite number of trials, an appropriate value of $n$ must be chosen for each experiment.  
In the following experiments, we set $n=60$.

\subsection{Prediction accuracy}

\begin{figure}[t]
  \centering
  \includegraphics[width=.9\linewidth]{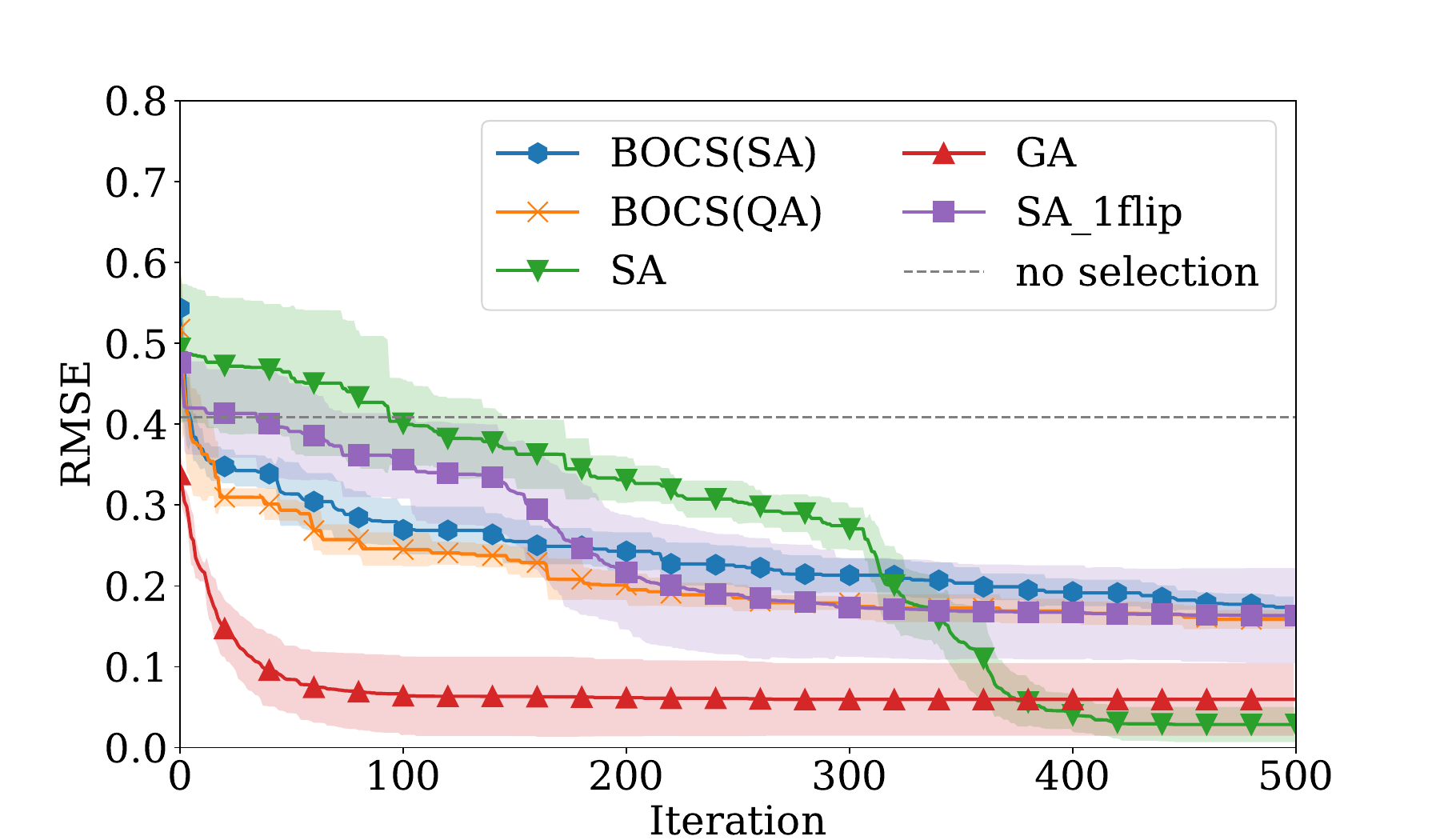}
    \caption{(Color online) RMSE trajectories of the compared wavelength-selection methods. The dashed gray line indicates the result without wavelength selection. The solid lines and shaded areas represent the mean values and standard deviations, respectively. Note that both GA and SA required 60 evaluations at each iteration. However, 1-flip SA, denoted as `SA\_1flip', evaluated $L(\bm x)$ only once at each iteration. Every method reduces validation loss after 500 iterations.}
  \label{fig_RMSE}
\end{figure}

Figure~\ref{fig_RMSE} compares the RMSE trajectories of the proposed method and the baseline methods.  
All wavelength-selection methods improved the prediction accuracy relative to the case without wavelength selection, whose RMSE was approximately 0.4.  
The measured sugar contents in the present dataset were approximately $20\,^\circ$Bx. 
Therefore, the final RMSE values below about $0.2\,^\circ$Bx correspond to an estimation error of roughly 1\% of the original physical scale, 
indicating that the proposed wavelength-selection approach achieves practically meaningful accuracy for real spectral data.
In the optimization of an acquisition function in BOCS, SA and QA showed no clear difference for the present dataset. 
This result suggests that, 
for the present problem size and wavelength-region resolution, the choice between SA and QA as the optimizer of an acquisition function was not the dominant factor determining the final prediction accuracy.
We focus on BOCS with SA in the following analysis.

The proposed method reached low RMSE values within fewer expensive PLS objective evaluations than the GA baseline.  
SA also improved the objective value but showed larger fluctuations and less reproducible selected subsets.
To ensure a consistent budget of objective function evaluations, we also tested the performance of random 1-bit flip SA. After 200 iterations, the random 1-bit flip SA achieved the same validation loss as BOCS-SA did.
One major drawback of this method is handling a temperature schedule, especially when the scale of $L(x)$ is unknown, while BOCS does not require explicit parameter tuning.

We note that the RMSE curve of the SA exhibits an abrupt drop in the later stage of the search. 
This behavior is attributed to the cooling schedule: around this stage the temperature becomes low enough that the acceptance probability for uphill moves is practically negligible, 
and the search effectively reduces to greedy descent. 
The sudden decrease is therefore not an intrinsic feature of the optimization but rather an artifact of the temperature schedule. 
This is consistent with the poor reproducibility of the wavelength subsets proposed by SA compared with BOCS, as discussed in the following sections.

\subsection{Robustness of selected wavelength regions}

\begin{figure}[t]
  \centering
  \includegraphics[width=.9\linewidth]{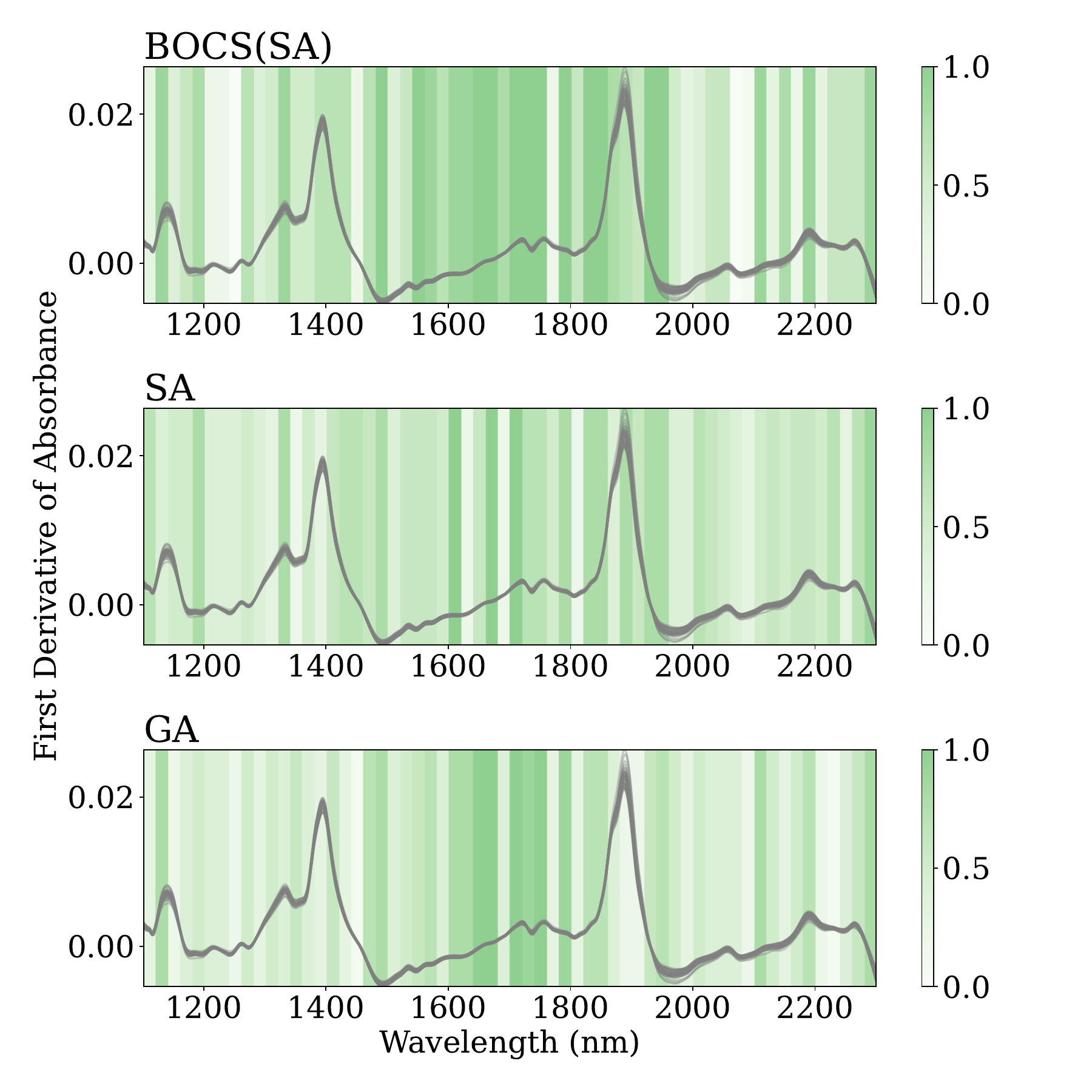}
    \caption{(Color online) Selected wavelength regions obtained by BOCS, the GA, and SA. The horizontal axis represents wavelength, and the color intensity represents the selection frequency, i.e., how often each region was selected over repeated runs. The solid lines indicate the first derivative of absorption spectra. BOCS-SA tends to select the range from \qtyrange{1500}{1980}{\nano\metre} more consistently than other two methods.}
  \label{fig_result_BOCS}
\end{figure}

Figure~\ref{fig_result_BOCS} shows representative selected wavelength regions obtained by BOCS, the GA, and SA.  
In each panel, 
the horizontal axis represents wavelength, and the color intensity represents the selection frequency of each region over repeated runs, that is, how often each wavelength region was chosen.  
BOCS repeatedly selected several specific wavelength regions, which appear as a small number of consistently darker bands, 
whereas the selected regions varied more broadly for the GA and SA.

The consistency of the selected subsets across repeated runs is quantified by two complementary metrics, listed in Table~\ref{tab:selected_wavelengths}: 
the mean Hamming distance and the per-bit Shannon entropy.  
The same table also lists the mean increase in RMSE caused by a one-spin removal, which summarizes the local robustness analysis discussed below.  
The mean Hamming distance measures how many wavelength regions differ, on average, between two independent runs; 
a smaller value indicates that repeated runs converge to more similar subsets.  
The per-bit Shannon entropy instead characterizes the selection uncertainty of each individual region.  
For each region, we estimate its selection probability over repeated runs and compute the associated binary entropy, averaged over all regions and normalized to lie in $[0,1]$.  
A value close to zero means that each region is chosen almost deterministically, either consistently selected or consistently excluded, 
whereas a value close to one means that the selection of each region fluctuates like an unbiased coin flip.

BOCS gave a mean Hamming distance of $19.2\pm0.4$ and a per-bit Shannon entropy of $0.61\pm0.02$, both of which were smaller than the corresponding values obtained by the GA and SA.  
The smaller Hamming distance shows that BOCS returns subsets that differ less from run to run, while the smaller Shannon entropy shows that this consistency holds at the level of individual wavelength regions, whose selection is more deterministic than for the baseline methods.  
Because the two metrics capture consistency at the subset level and at the per-region level, respectively, their agreement provides stronger evidence that the reproducibility of BOCS is a robust property rather than an artifact of a particular distance measure.  
Such consistency is useful when the selected spectral regions are to be inspected by domain experts, because it reduces ambiguity caused by stochastic optimization and allows the repeatedly chosen regions to be interpreted with greater confidence.

\begin{table*}[t]
  \centering
  \caption{Run-to-run consistency and local robustness of selected wavelength regions.  The first two rows quantify selection consistency across repeated runs.  The last row gives the mean increase in RMSE caused by removing one selected region at a time, corresponding to the one-spin-removal analysis in Fig.~\ref{fig_robustness}.  Smaller values indicate more consistent and more locally robust selection.  Uncertainties denote standard errors.}
  \label{tab:selected_wavelengths}
  \begin{tabular}{lccc}
    \hline
      & BOCS (SA) & GA & SA \\
    \hline
    Mean Hamming distance & $19.2\pm0.4$ & $25.6\pm0.5$ & $27.1\pm0.5$ \\
    Shannon entropy / \# bits & $0.61\pm0.02$ & $0.80\pm0.02$ & $0.84\pm0.02$ \\
    Mean $\Delta$RMSE under one-spin removal & $0.044\pm0.003$ & $0.147\pm0.005$ & $0.219\pm0.007$ \\
    \hline
  \end{tabular}
\end{table*}

\begin{figure}[t]
  \centering
  \includegraphics[width=.9\linewidth]{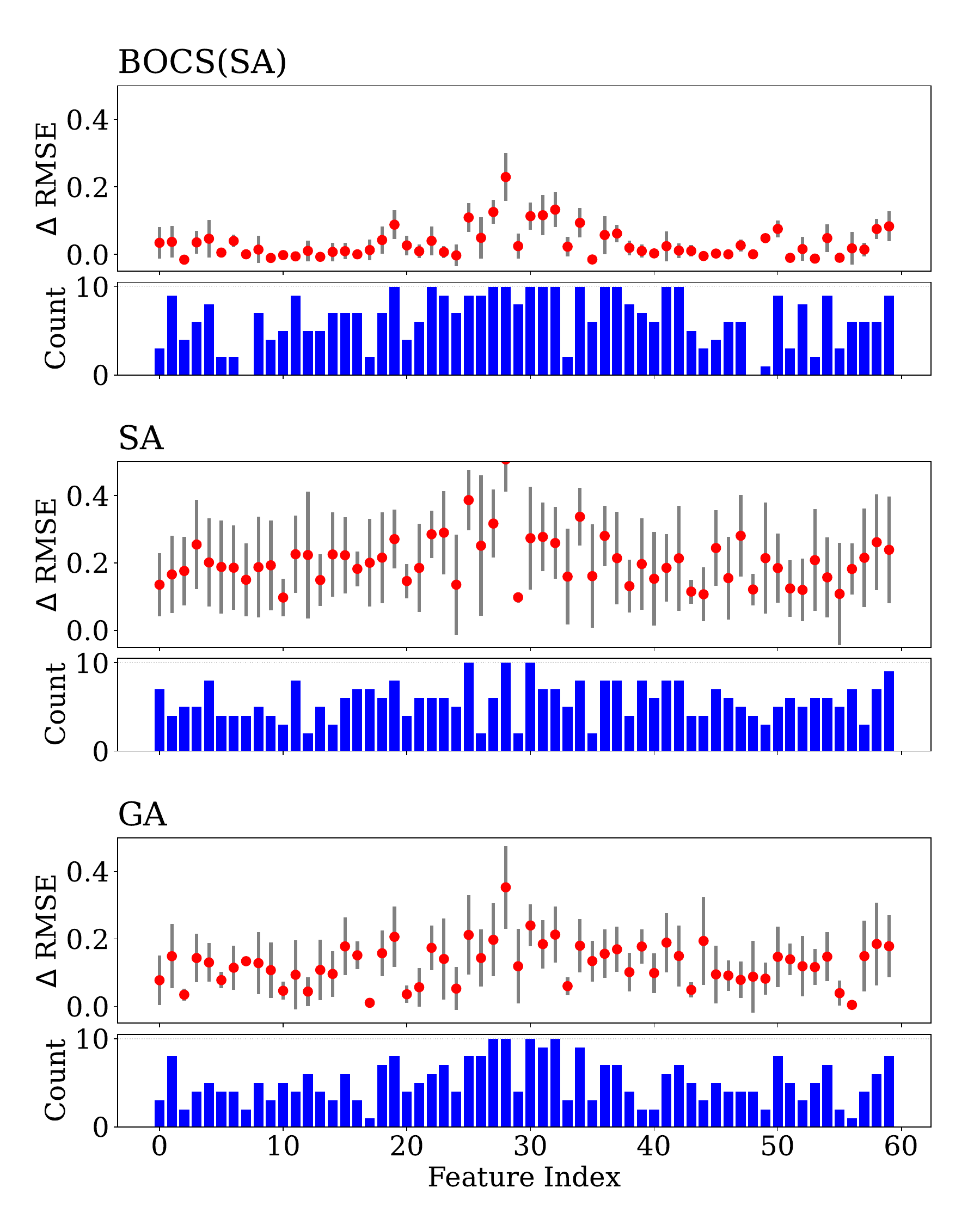}
    \caption{(Color online) Local robustness of the selected wavelength subsets. For each method, the upper panel shows the change in RMSE when one selected component is removed, and the lower panel shows the selection count of each component over repeated runs. Note that the subtraction of RMSEs was performed only at each selected small wavelength region ($x_i = 1$) but not was at unselected one ($x_i = 0$). The results from BOCS-SA demonstrates smaller mean RMSE differences and its more consisitent selection from run to run.}
  \label{fig_robustness}
\end{figure}

Beyond the run-to-run consistency discussed above, 
we next examined the local stability of the selected subsets, namely their robustness to small changes in the selected regions themselves.  
For each method, we removed one selected wavelength region at a time from a representative selected subset and measured the resulting increase in RMSE.  
The results are shown in Fig.~\ref{fig_robustness}, and the corresponding mean increases in RMSE are summarized in Table~\ref{tab:selected_wavelengths}.

Compared with SA and the GA, 
BOCS had only a small number of regions whose removal caused a large change in RMSE, while the removal of most regions left the RMSE almost unchanged.  
Quantitatively, the mean increase in RMSE under one-spin removal was $0.044\pm0.003$ for BOCS, whereas it was $0.147\pm0.005$ for the GA and $0.219\pm0.007$ for SA.  
Thus, on average, removing one selected wavelength region from the BOCS solution caused a substantially smaller degradation in prediction accuracy than in the baseline methods.  
This result suggests that 
the subset selected by BOCS is locally stable: its neighboring subsets, obtained by a single one-bit removal, retain comparable prediction performance in most cases.  
At the same time, 
the few regions that do lead to a large increase in RMSE can be interpreted as particularly important for prediction, since the accuracy of the model depends critically on their inclusion.  
Notably, these regions coincide with those repeatedly selected across runs in Fig.~\ref{fig_result_BOCS}, so that the run-to-run consistency and the local importance of individual regions reinforce each other.

This tendency of BOCS to select flat solutions, whose neighbors also achieve low error, can be understood from the structure of the proposed method.  
The sparse quadratic surrogate model, regularized by the horseshoe prior, captures a small set of informative regions together with their pairwise interactions, rather than fitting isolated low-error configurations of the objective.  
Because Thompson sampling repeatedly proposes and evaluates candidates around the regions supported by this surrogate, the search concentrates on broad basins in which neighboring subsets also yield low RMSE, instead of narrow, isolated minima.  
The smaller mean $\Delta$RMSE under one-spin removal in Table~\ref{tab:selected_wavelengths} is consistent with this interpretation, indicating that the BOCS solution is surrounded by neighboring subsets with similar predictive performance.  
In contrast, SA and the GA optimize the expensive objective directly and are therefore more prone to settling in isolated, overfitted solutions, which is consistent with their larger average RMSE increases and the broader spread of RMSE changes in Fig.~\ref{fig_robustness}.

\section{Conclusion and Future Work}
\label{sec_con}




We proposed a robust wavelength-selection method for near-infrared spectral prediction based on combinatorial Bayesian optimization.
The wavelength-selection problem was formulated as a binary black-box optimization problem, where the objective was the cross-validated RMSE of partial least squares regression.
The proposed method used a sparse quadratic surrogate model regularized by a horseshoe prior and Thompson sampling to propose informative wavelength subsets, and the resulting QUBO problem was solved by SA or QA.
In this framework, the expensive PLS objective is evaluated only once per proposed subset, so that the costly evaluations are used sample-efficiently.

In experiments on prune near-infrared spectra, the proposed method achieved accurate sugar content estimation with fewer objective evaluations than the genetic-algorithm baseline, while SA and QA yielded comparable performance in Thompson sampling for the present problem size.
Beyond prediction accuracy, we assessed the selected subsets from two complementary viewpoints.
First, in terms of run-to-run consistency, BOCS achieved the smallest mean Hamming distance and per-bit Shannon entropy among the compared methods, indicating that its reproducibility holds both at the subset level and at the level of individual wavelength regions.
Second, in terms of local robustness, the subsets selected by BOCS remained stable under most one-bit removals, and the few regions whose removal degraded the accuracy coincided with those repeatedly selected across runs.
These two properties reinforce each other and suggest that BOCS tends to identify broad, stable basins rather than isolated, overfitted solutions, in contrast to the direct optimization performed by SA and the GA.

Taken together, these findings indicate that BOCS is a useful framework for feature selection in high-dimensional spectral data, particularly when objective evaluations are expensive and the reproducibility of the selected features is important.
Because the selected regions are consistent and locally robust, they can be presented to domain experts with greater confidence, which is a practical advantage when the goal is not only accurate prediction but also interpretable identification of informative spectral regions.
Moreover, since the method relies only on a QUBO optimization step, it is naturally compatible with QA hardware and is expected to benefit from future improvements in annealer scale and coherence, especially as the number of wavelength regions and the density of pairwise interactions grow.

Several directions remain for future work.
A more reliable estimate of the generalization performance may require nested validation with an independent test set, which may also clarify the extent of any selection bias in the reported errors.
The resolution of the wavelength regions, controlled by $n$, was chosen from the prediction results in the present study. Its automatic determination, therefore, may make the method fully data-driven.
Chemical validation of the selected regions, for example by relating them to known absorption bands associated with sugar content, may further strengthen the interpretability of the results.
Finally, the same framework may be applicable to other spectral data and to high-dimensional feature-selection problems outside spectroscopy, wherever expensive black-box objectives and a demand for reproducible feature subsets coincide.

\begin{acknowledgment}
The authors thank Mr. Kosaka and Mr. Kitade from DIC Corporation for their support and discussions during this study.
\end{acknowledgment}


\bibliographystyle{jpsj-custom}
\bibliography{main}

\end{document}